\documentclass[]{interact}

\usepackage{array, blkarray}

\usepackage{arydshln}
\usepackage{verbatimbox}
\usepackage{graphicx}
\usepackage{url}
\usepackage{hyperref}
\usepackage{amsmath,amsfonts}
\usepackage[linesnumbered,lined,boxed,commentsnumbered,ruled]{algorithm2e}
\usepackage{textcomp}
\usepackage{multirow}
\usepackage{multicol}
\usepackage{color}
\usepackage[round]{natbib}
\usepackage{caption}
\usepackage{setspace}
\usepackage{comment}
\usepackage{threeparttable}
\usepackage{booktabs}
\usepackage{tikz}
\usetikzlibrary{arrows,positioning}
\usepackage{lipsum, adjustbox}

\usepackage{siunitx}
\usepackage{subcaption}

\newcommand{\tcolR}{\textcolor{black}}

\begin{document}


\title{Time Series Clustering for Grouping Products Based on Price and Sales Patterns}

\author{
\name{Aysun Bozanta\textsuperscript{a}, Sean Berry\textsuperscript{a}, Mucahit Cevik\textsuperscript{a}\thanks{CONTACT M. Cevik Author. Email: mcevik@ryerson.ca}, Beste Bulut\textsuperscript{b}, Deniz Yigit\textsuperscript{b}, Fahrettin F. Gonen\textsuperscript{b}, and Ay\c{s}e Ba\c{s}ar\textsuperscript{a} }
\affil{\textsuperscript{a}Data Science Lab at Ryerson University, Toronto, Canada; \textsuperscript{b}Getir Perakende Lojistik A.S., Istanbul, Turkey}
}

\maketitle

\begin{abstract}
Developing technology and changing lifestyles have made online grocery delivery applications an indispensable part of urban life. 
Since the beginning of the COVID-19 pandemic, the demand for such applications has dramatically increased, creating new competitors that disrupt the market. 
An increasing level of competition might prompt companies to frequently restructure their marketing and product pricing strategies. 
Therefore, identifying the change patterns in product prices and sales volumes would provide a competitive advantage for the companies in the marketplace. 
In this paper, we investigate alternative clustering methodologies to group the products based on the price patterns and sales volumes.
We propose a novel distance metric that takes into account how product prices and sales move together rather than calculating the distance using numerical values. 
We compare our approach with traditional clustering algorithms, which typically rely on generic distance metrics such as Euclidean distance, and image clustering approaches that aim to group data by capturing its visual patterns. 
We evaluate the performances of different clustering algorithms using our custom evaluation metric as well as Calinski Harabasz and Davies Bouldin indices, which are commonly used internal validity metrics. 
We conduct our numerical study using a propriety price dataset from an online food and grocery delivery company, and the publicly available Favorita sales dataset.
We find that our proposed clustering approach and image clustering both perform well for finding the products with similar price and sales patterns within large datasets.
\end{abstract}


\section{Introduction}
E-commerce is one of the fastest-growing channels for global online grocery delivery, recording a \$198.5 billion global market in 2020, and is projected to reach a market size of \$550.7 billion by 2027 \citep{researchandmarkets2021}. 
The conditions created during the pandemic, the developing technology, and the changing lifestyles all point to the need for such online platforms. 
Thus, the number of companies who have recently participated in the market has rapidly increased. 
With the growth of the market and the entry of different players, dynamic product pricing and sales strategies have become an important tool for the companies. 
The strategic and competitive product pricing and sales might move companies one step ahead of their competitors, and help them gain a competitive advantage. 

Identifying the products with similar price and sales patterns over a given time period might provide a better understanding of the market, and create opportunities for better marketing and pricing strategies in the long run.
In this regard, time series clustering can help to obtain the products whose sales volumes and prices move similarly over a given time period. 
By examining the generated clusters, we may also gain insights into the time periods in which certain products have been discounted or got a price hike.

Existing clustering algorithms commonly use distance metrics that measure the numerical distance between the data points. 
These metrics are not capable of capturing the simultaneous change in the time series patterns. 
That is, even if two time series move in opposite directions in a given time period, if they are numerically close to each other, these metrics will find the distance between these two time series to be small, and disregard the movement patterns of the time series. 
In terms of pricing, this might prevent capturing the movement patterns that are useful for obtaining the price change points and directions, and only enable putting time series with close prices in the same clusters. 
On the other hand, capturing the price change patterns implies regardless of the value, if the prices of the products move together in the same periods (i.e., price drops or price hikes), then they should fall in the same cluster. 

In this study, we investigate clustering methods to group similar grocery products in terms of price and sales volume change patterns. 
For this purpose, we develop a custom distance metric that measures the distance between two time series based on their movement patterns, which is then used within a hierarchical clustering algorithm to obtain the product groupings.
In addition, we apply image clustering algorithms that aim to group similar data instances based on their visual time series patterns. 
We also experiment with standard clustering approaches (e.g., $k$-means clustering) by using generic distance metrics (e.g., Euclidean distance).
One of the challenging tasks in clustering analysis is to measure the performance of the generated clusters. 
In particular, when there are no cluster labels (i.e., the ground truth), only the internal validity metrics can be used to evaluate the quality of the clusters, which rely on generic distance metrics. 
\tcolR{However, since the generic metrics such as Euclidean and Manhattan distances only consider the numerical closeness of the data points within the clusters,
they are not suitable for our problem, which prompted us to develop a custom evaluation metric. 
}

The contribution of our study can be summarized as follows.
\begin{itemize}
    \item Our study provides an application of time series clustering for grocery product grouping based on price and sales volume history, which is an important practical problem for marketing and sales. 

    \item To the best of our knowledge, image clustering has not been previously used for the task of clustering products based on price and sales volumes.
    We find this approach to be effective in capturing common patterns between different time series, and hence grouping the grocery products using the price and sales histories.

    \item We propose a custom distance metric that measures the distance between two time series based on their movement patterns.
    This metric can also be used for time series clustering tasks other than grocery product grouping.
    We also develop a custom evaluation metric to evaluate the performance of the clustering algorithms, which scores the clusters using our custom distance metric.
    
\end{itemize}

The remainder of the paper is organized as follows. 
Section~\ref{sec:litrev} provides an overview of clustering approaches with a particular focus on product clustering with different techniques. 
Section~\ref{sec:method} provides a summary of the methodologies used, dataset characteristics, and a detailed discussion of our preprocessing, algorithms, and evaluation metrics.
In Section~\ref{sec:results}, we present the results of our numerical study, which includes an illustrative example of how our custom distance metric works, and a performance comparison of clustering algorithms with different distance metrics by using various evaluation criteria.
Lastly, Section~\ref{sec:conc} provides concluding remarks and future research directions.

\section{Literature Review}\label{sec:litrev}
Time series clustering has been used in diverse scientific domains to discover hidden patterns from complex and massive datasets. 
Clustering methodology in time series analysis can be examined with respect to the data representation, distance metric, and clustering algorithm \citep{aghabozorgi2015time}. 
The application of data transformation techniques on the original data may yield better results in clustering algorithms. 
Commonly used data representation techniques in the literature are Discrete Fourier Transformation (DFT), Singular Value Decomposition (SVD), and Discrete Wavelet Transform (DWT) \citep{aghabozorgi2015time}. 
Another component of clustering is the distance metric that is used within the clustering algorithm. 
Well-known distance metrics include Euclidean distance, Dynamic Time Warping (DTW), and Longest Common Subsequence (LCSS) \citep{aghabozorgi2015time}. 
The clustering approaches are mainly classified into six groups, namely, partitioning, hierarchical, grid-based, model-based, density-based clustering, and multi-step clustering. 
On the other hand, for time series analysis, mostly partitioning and hierarchical clustering techniques have been employed. 

\citet{alvarez2010} applied time series forecasting on the electricity demand dataset. 
Before the time series forecasting, they used $k$-means clustering to group and label the time series data. 
They evaluated the performance of the clustering algorithm by three well-known internal validity indices, namely, silhouette index, Davies-Bouldin (DB) index, and the Dunn (DU) index. 
\citet{bandara2020forecasting} proposed a novel forecasting model which can be used with different types of RNN models on subgroups of similar time series. 
Their clustering technique is feature-based, which, instead of capturing similarity of point values using a distance metric, employs sets of global features obtained from a time series to summarize and describe the salient information of the time series. 
They noted that feature-based approaches can be more interpretable and more resilient to missing and noisy data. 
\citet{javed2020benchmark} used the time series datasets for the University of California Riverside (UCR) archive and conducted a benchmark time series clustering study. 
They combined the $k$-means, fuzzy $c$-means, agglomerative clustering, and density-based clustering algorithms with the Euclidean, DTW, and shape-based distance metrics, 
and compared alternative clustering approaches using the rand index.

Unlike other clustering approaches, image clustering determines the visual patterns of the data and groups similar patterns together.
This method consists of two components: feature learning and clustering.
For the feature learning phase, deep learning techniques have been most commonly used in the literature \citep{wu2019deep, ren2020deep}, 
whereas, for the clustering phase, $k$-means clustering is typically employed \citep{caron2018deep}. 
\citet{ahmed2015recent} applied different image clustering algorithms on 14 benchmark image datasets and observed that the performance of image clustering algorithms depends mostly on the distribution of the image datasets. 
The application of image clustering techniques on the time series data is very limited, and they have been mostly used on the satellite image time series datasets \citep{khiali2019detection, lampert2019constrained}.

Time series modeling is frequently used in sales and marketing, especially for forecasting future sales and demand~\citep{Pavlyshenko_2019}. 
Accurate forecasting of future sales and demand allows companies to leverage information from historical data strategically to increase future revenue~\citep{Leonard_2000}. 
There is an extensive literature on time series models for sales and demand forecasting.
\citet{Abbasimehr_Shabani_Yousefi_2020} developed multi-layer LSTM networks and automated hyperparameter tuning approaches for demand forecasting, and showed their to be effective when compared to some other popular models such as ARIMA and support vector machines. 
\citet{Pavlyshenko_2019} compared different machine learning approaches for sales forecasting problem, and found that a model stacking approach could reliably improve model performance to a significant degree.
For more information on time series models for sales and demand forecasting, we refer the reader to a recent review by \citet{fildes2019retail}.
Despite the popularity of time-series forecasting over sales and demand datasets, there has been very little research done in the area of time series clustering for these types of data. 

Although the time series clustering approaches have been considered in various domains such as climate \citep{awad2018enhanced}, biology \citep{mcdowell2018clustering}, finance \citep{takahashi2019modeling}, environment and urban planning \citep{fu2019characterizing}, the applications of time series clustering techniques on product clustering is very limited. 
In addition, image clustering approaches have not been commonly used on time series data, and to the best of our knowledge, previous studies did not consider clustering approaches for product grouping based on price and sales volume histories.

\section{Methodology}\label{sec:method}
\begin{figure*}[!ht]
    \centering
    \includegraphics[width=0.95\textwidth]{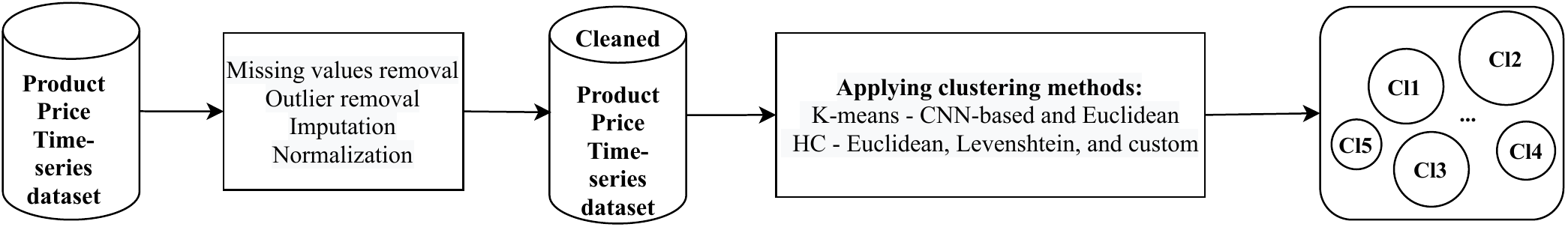}
    \caption{Overall methodology for grouping products based on their price histories, which are represented as distinct time series for each product (similar steps are followed for product clustering based on sales volume).}
    \label{fig:method}
\end{figure*}

In this section, we discuss the datasets, our proposed distance metric for time series data, the image clustering approach, and the evaluation metrics for measuring clustering performance. 
Figure~\ref{fig:method} presents the overall methodology of this study. 

\subsection{Datasets}
\tcolR{We consider two datasets in our numerical analysis.} 
The first dataset consists of a collection of price values over 3,285 products for a period of $n=570$ days.
This dataset is obtained from 
XYZ company\footnote{Blinded for peer review}
an online food and grocery delivery company that originated in Turkey and recently expanded its operations to the United Kingdom and the Netherlands. 
Daily price values for the products are available as numerical values in the dataset.
\tcolR{The second dataset is the publicly available Favorita sales data\footnote{https://www.kaggle.com/c/favorita-grocery-sales-forecasting/data}. 
The Favorita dataset was provided by Corporacion Favorita, an Ecuadorian-based grocery retailer, with the aim of predicting future sales of their products. 
Favorita sales data contains 59,038,132 observations over 54 stores, taken between January 1st 2016 and August 15th 2017. Each observation includes a date, a store number, item number, the unit sales for the day, as well as whether the item had a promotional price that day. 
After aggregating the distinct products and stores, we end up with 172,130 distinct product store pairs.
}

We apply a series of data preprocessing steps for both datasets.
We format the raw data as time series by aggregating all the price\tcolR{/sales} values for each product over time.
Real-world time series data is often rife with aberrations.
We remove any product with over 80\% of their prices\tcolR{/sales} unavailable. 
For the remaining products, many of which have missing price values, we fill the missing prices with the most recently available price for that product. 
\tcolR{For Favorita dataset, we fill the missing sales with the mean values.}

\begin{figure}[!ht]
    \centering
    \subfloat[Price data \label{fig:sample_price}]{\includegraphics[width=0.445\textwidth]{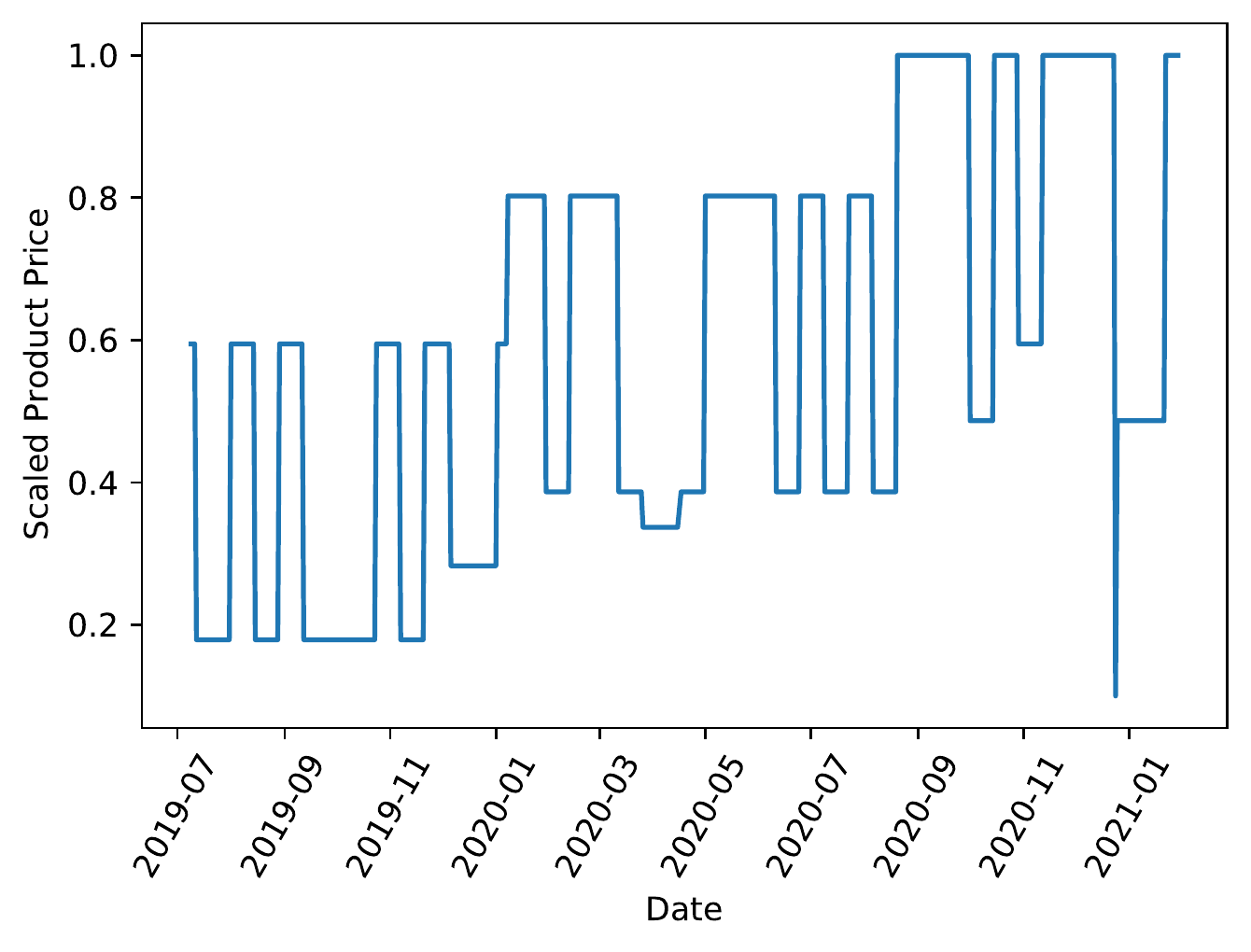}}\\[0.3em]
    \subfloat[Sales data \label{fig:sample_sale}]{\includegraphics[width=0.825\textwidth]{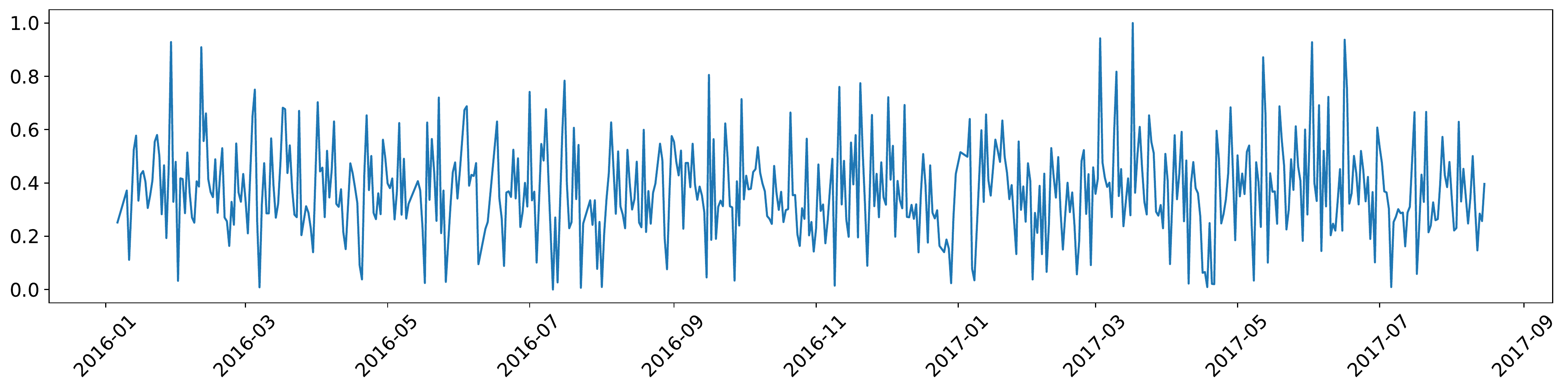}}
    \caption{Time series samples.}
    \label{fig:sample_TS}
\end{figure}

We next use a min-max scaler to scale the prices\tcolR{/sales} of each product\tcolR{/product-store pair} to be between 0.1 and 1. 
This approach is helpful since we are interested in relative price\tcolR{/sales} fluctuations, and products with different prices\tcolR{/sales} but similar fluctuations are aimed to be grouped together. 
After proper cleanup, scaling, and exploratory data analysis, we note that many of the time series are not suitable to be grouped in any clusters. 
We examine the pairwise distances between time series and remove those that are not likely to be clustered together with any other time series in the dataset.
Sample time series for price and sales data is provided in Figure~\ref{fig:sample_price} and Figure~\ref{fig:sample_sale}, respectively.
We note that, while price data shows little fluctuations over time, as expected, sales values are highly variable.

\tcolR{At the end of the preprocessing phases, 1,160 product price time series remained in the price dataset, while 134 product-store sales time series remained in the sales dataset. 
Those are then discretized for our custom metric as well as any string-based clustering method that focuses on movement patterns rather than the actual numerical values.} 
Specifically, we use the following function that considers equal distances between the normalized values, which are identified based on preliminary analysis.
\begin{align}
\label{eq:customDistance}
z(x) = \begin{cases} 
      A & x < 0.29 \\
      B & 0.29\leq x < 0.47 \\
      C& 0.47\leq x < 0.65\\
      D &  0.65\leq x < 0.83 \\
      E   &  0.83 \leq x
\end{cases}
\end{align}

\subsection{Distance Metrics}
We employ three distance metrics, namely, Euclidean Distance, Levenshtein Distance, and Movement Pattern-based Distance, which we briefly summarize next.
\subsubsection{Euclidean Distance} 
Let $p$ and $q$ be two product time series from our dataset. Then the Euclidean distance between these products is given by $D(p,q) = \sqrt{\sum_{i=1}^n (q_i - p_i)^2}$.

\subsubsection{Levenshtein Distance} This is a string metric that can be simply defined as the minimum number of single-character edits required to change one word into the other. 
In our case, this corresponds to the number of single-character edits required to change the discretized prices/sales of one product into the other. 
Let $p$ and $q$ be two product time series from our dataset. 
Then the Levenshtein distance between these products is given by using the below recursive function:
\begin{align*}
D(p,q) = \begin{cases} 
      |p| & \text{if}\ |q|=0 \\[0.2em]
     |q| & \text{if}\ |p|=0 \\[0.2em]
      D(p[1:],q[1:]) &  \text{if}\ p[0] = q[0] \\[0.2em] 
      1 + \min
    \begin{cases}
       D(p[1:],q[1:]) &\\[0.2em]
       D(p,q[1:]) &\\[0.2em]
       D(p[1:],q) & \\[0.2em]
    \end{cases} & \text{otherwise}
   \end{cases}
\end{align*}

\subsubsection{Movement Pattern-based Distance}
We propose a custom distance metric, the movement pattern-based distance (MPBD), which takes into account the direction of change in a time series.
MPBD is developed by iterating through two time series simultaneously and following the below rules. 
\begin{itemize}
    \item If both time series have constant values at the same time, no distance is added. 
    \item If both time series move in the same direction, and the same magnitude at the same time no distance is added. 
    \item If both time series move in the same direction, but at a different magnitude, absolute value of the difference between the magnitudes is added. 
    \item If both time series move in the opposite directions, a distance value corresponding to the absolute value of the difference between the magnitudes multiplied by some weight (e.g., $\omega=2$) is added.
\end{itemize}
In this manner, we aim to capture products (i.e., time series) that move similarly even with different price points or relative price points.
Let $d_p = p[0] - p[1] $ for product $p$ in the dataset. 
A naive recursive definition of the distance function for two products $p$ and $q$ can be obtained as follows. 
\begin{align*}
\begin{cases} 
0 & \text{if } |p|=|q|=1\\
0 + D(p[1:],q[1:]) & \text{if } d_p \And d_q = 0 \\
0 + D(p[1:],q[1:]) & \text{if } d_p = d_q  \\
|d_p-d_q| + D(p[1:],q[1:]) & \text{if } d_p \ne d_q \And \\ & sign(d_p) = sign(d_q) \\
\omega \cdot |d_p-d_q| + D(p[1:],q[1:]) & \text{if }  sign(d_p) \ne sign(d_q) \\
   \end{cases}
\end{align*}

\subsection{Clustering Algorithms}
We consider two traditional clustering algorithms ($k$-means and hierarchical clustering) as well as image clustering for grouping the product time series.
\subsubsection{$k$-means Clustering} 
$k$-means is a popular clustering algorithm, which aims to minimize the within-cluster sum of squares (WCSS) measure, that is,
\begin{equation}\label{wsse}
   WCSS = \sum_{i=1}^k \sum_{j=1}^n (x_{ij} - \mu_i)^2
\end{equation}  
where $x_{ij}$ is item $j$ in cluster $i$ and $\mu_i$ is the corresponding cluster centre of cluster $i$. 
Items are retroactively assigned to their nearest cluster center, and then the cluster centers are updated until the items in the clusters no longer change. 
    
\subsubsection{Hierarchical Clustering}
Hierarchical clustering requires either starting with all items in one cluster and then breaking that cluster into smaller sub-clusters (divisive hierarchical), or the exact opposite, i.e., starting with all items in their own cluster and joining clusters based on some criteria (agglomerative hierarchical). 
In our implementations, we use an agglomerative hierarchical approach with Ward's linkage criteria.

\subsubsection{Image Clustering}
Image clustering is an increasingly popular research area with many potential applications. 
We consider image clustering for time series data obtained over product price/sale histories.
Specifically, for each product in the dataset, we create an image illustrating the graph of the numerical values over time. 
We transform these images into array representations in three-dimensional space.
We then use the VGG16 convolutional neural network (CNN) to extract feature vectors.
Finally, we apply the $k$-means clustering algorithm on the feature vectors of the time series images.

\subsection{Evaluation Metrics}
Since our dataset does not include ground truth labels (i.e., actual cluster ids), we rely on internal indices for clustering performance assessment, which is briefly summarized below.

\subsubsection{Calinski Harabasz (CH) Index} 
This index is defined as the ratio of the sum of between-clusters dispersion and inter-cluster dispersion. 
Thus a higher score implies that the clusters are very distinct from each other yet the cluster contents are dense. 
We define the between cluster sum of squares as follows.
\begin{align}\label{bcss}
BCSS = \sum_{i=1}^k(\mu_i-\mu)^2 
\end{align}
where $\mu_i$ is the cluster centre for cluster $i$ and $\mu$ is the sample mean for the whole dataset. 
Following from Eqn.~\ref{wsse} and Eqn.~\ref{bcss}, the CH index can be obtained as $CH = \frac{WCSS}{BCSS}$. 

\subsubsection{Davies Bouldin (DB) Index} 
This index is another popular internal validity metric, which aims to measure the average similarity of each cluster with a cluster most similar to it. 
A lower DB value is an indicator that the clusters are well separated. 
The calculation of the DB index is most easily represented in multiple steps as follows.
\begin{itemize}
    \item First, we calculate the intra-cluster dispersion for each cluster  with the formula $S_i = \sqrt{\frac{1}{n_i} \sum_{j=1}^{n_i} (x_{ij} - \mu_i)^2 }$, where $n_i$ is the number of items in cluster $i$  and $\mu_i$ is the cluster center of cluster $i$.
    
    \item Next, we calculate the distance between cluster centers for all clusters with the formula: $M_{i,j} = \sqrt{(\mu_i -\mu_j)^2 }$
    
    \item The similarity measure between two clusters $i$ and $j$ is thus $R_{i,j} = \frac{S_i+S_j}{M_{i,j}}$
    
    \item DB index is then given by $DB = \frac{1}{k}\sum_{i=1}^k \max_j(R_{i,j})$.
    
\end{itemize}

\subsubsection{Movement Pattern-based Index}
Let $D_{cust}(x,j)$ be the distance between two products $x$ and $j$ as defined by our custom distance metric, MPBD. 
Then a custom evaluation metric, which we refer to as movement pattern-based index (MPBI) can be obtained as follows
\begin{align*}
MPBI = \frac{1}{k}\sum_{c=1}^k \frac{1}{n} \Big[ \sum_{i=0}^{n-1} \sum_{j=i+1}^n D_{cust}(x_{ci},x_{cj}) \Big]
\end{align*}
where $x_{{ci}}$ denotes the $i$th product in cluster $c$, $n$ denotes the number of data points in a given cluster, and $k$ is the number of clusters. 
Specifically, this measure is the sum of the clusters' pairwise distances divided by the number of products in the same cluster and averaged over all the clusters. 
Therefore a lower MPBI value suggests low pairwise distances, which in turn suggest that the clustered items are close. 
Additionally, this index allows us to use the discretized data exclusively without the need for a numeric dataset.

\section{Results}\label{sec:results}
In this section, first, the working principle of MPBD is presented in an illustrative example. 
Then, the experimental results comparing the performances of the combination of two clustering algorithms with four different distance metrics are provided for a clustering task involving product price \tcolR{and product sales} time series. 
Lastly, we provide cluster profile results to show which products tend to be grouped together based on their price and sales histories.

\subsection{Illustrative Example}
\begin{figure}[!ht]
    \centering
    \subfloat[scenario1]{
    \includegraphics[width=0.495\textwidth]{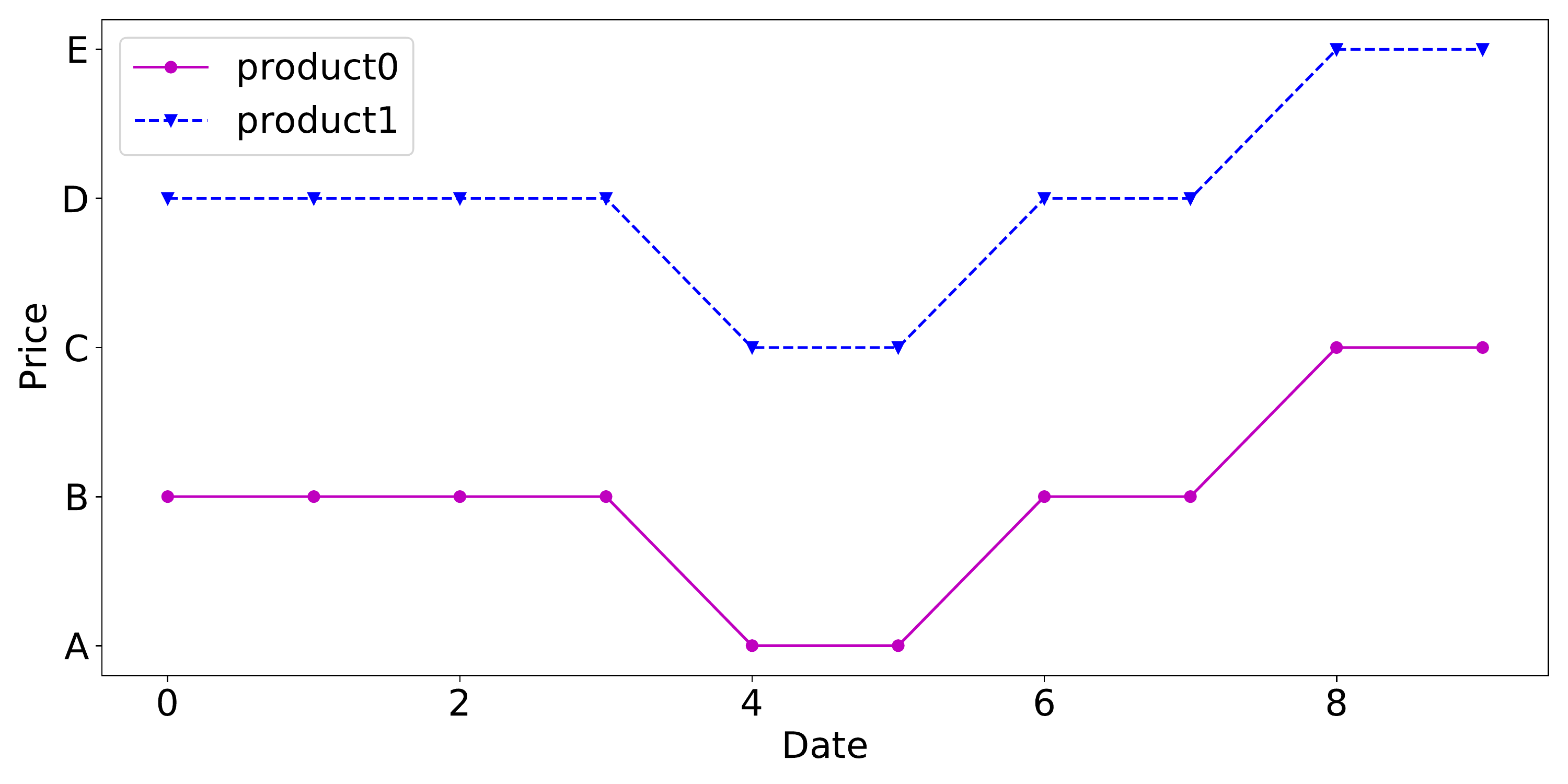}}
    \subfloat[scenario2]{
    \includegraphics[width=0.495\textwidth]{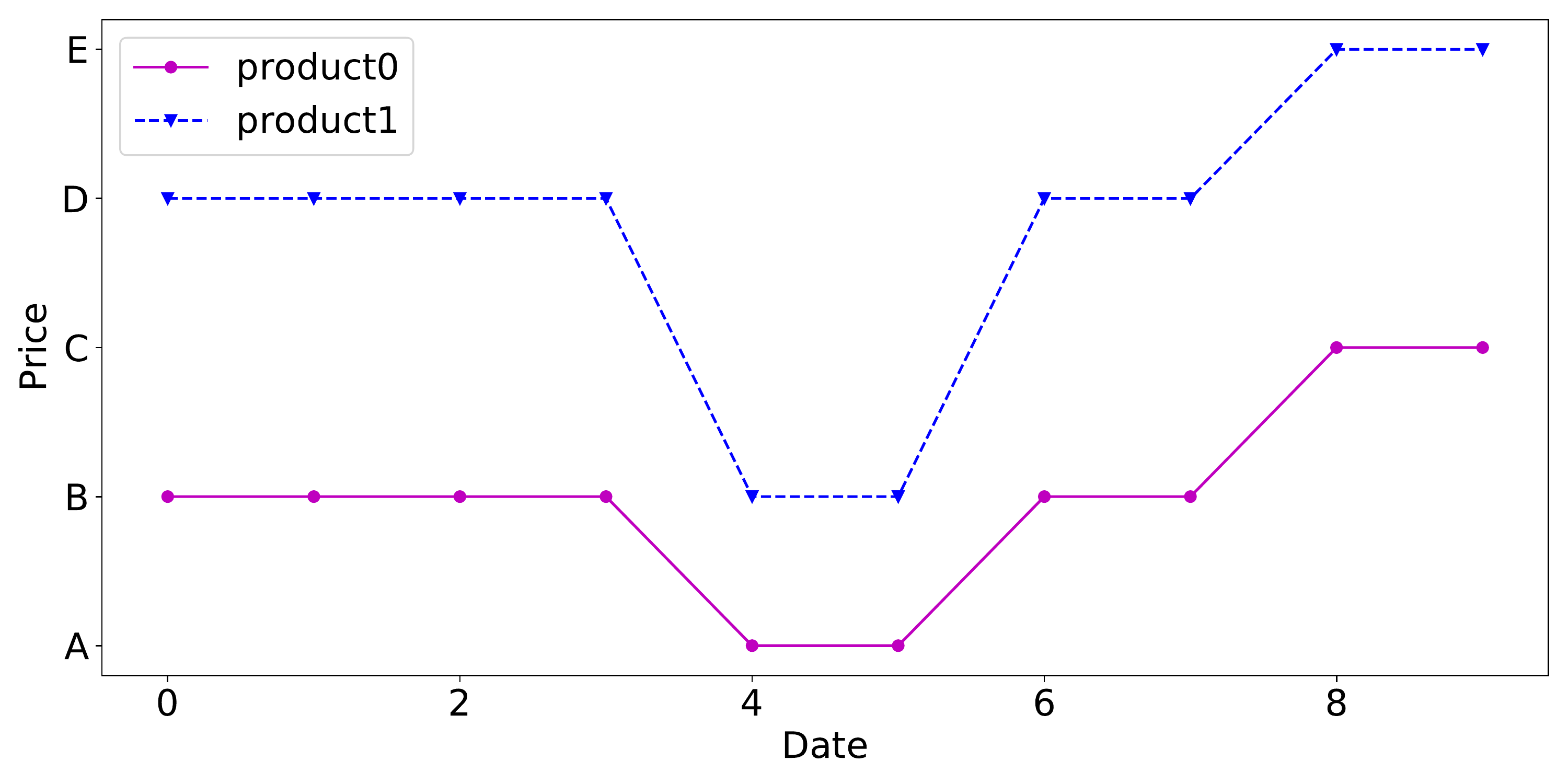}}\\
    \subfloat[scenario3]{
    \includegraphics[width=0.495\textwidth]{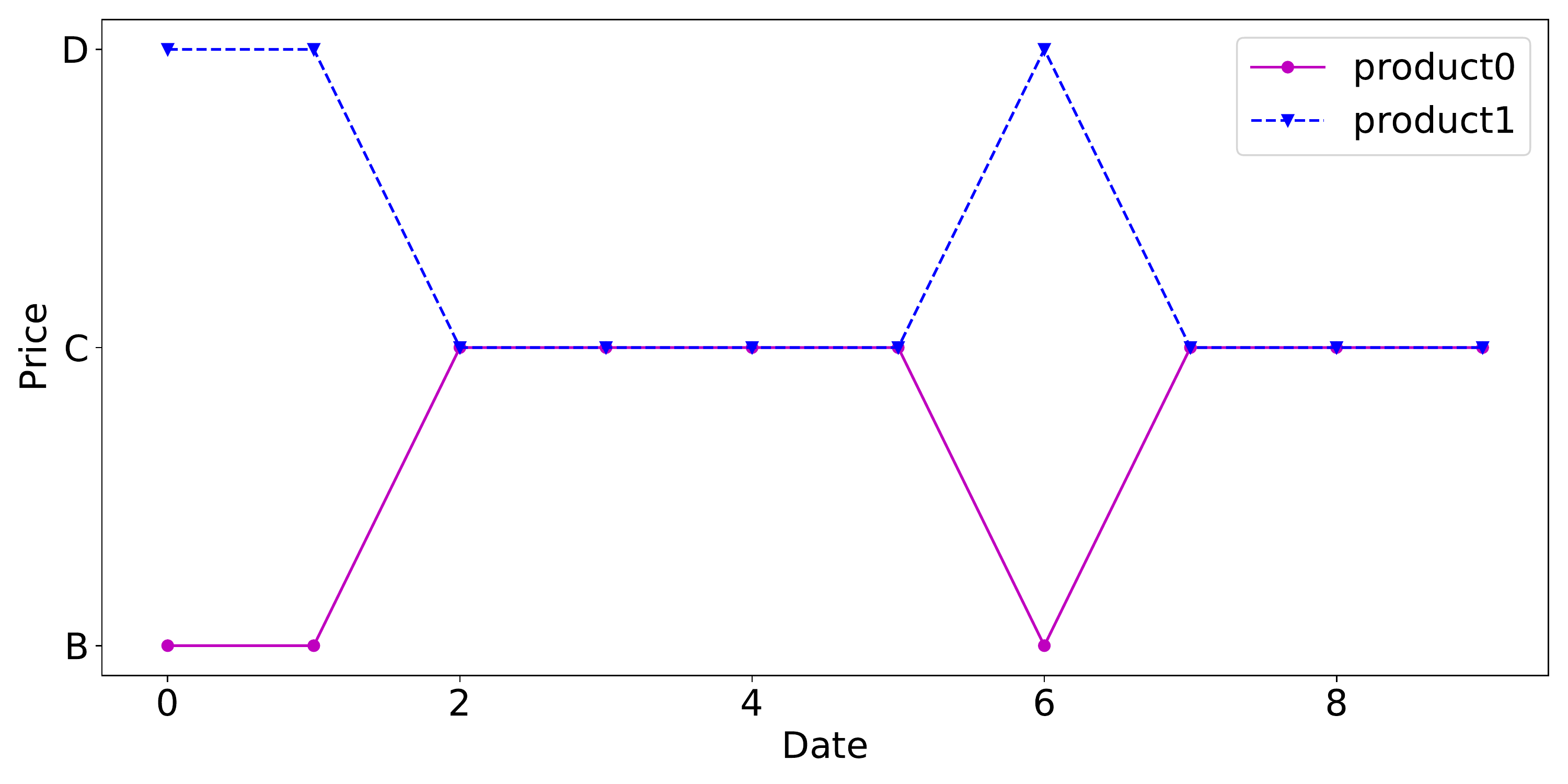}}

    \caption{ Time series price changes of two products in different scenarios}
    \label{fig:scenario}
\end{figure}

We created three scenarios over synthetic price time series to depict the difference between MPBD, Euclidean distance, and Levenshtein distance.  
Figure~\ref{fig:scenario} presents the price changes of two products over a ten-day period for three scenarios. 
Table~\ref{tab:scen_table} shows the normalized distance values between two time series representing product prices, calculated by using different distance metrics. 
In scenario 1, although the price levels of the two products are entirely different from each other, they show the same price change pattern. 
Since MPBD considers the price change similarity instead of the exact numerical closeness of two prices, its value becomes 0.0 for scenario 1. 
On the other hand, because each price value in each time step is different from each other for scenarios 1 and 2, Levenshtein distance values become 1.0, which is the maximum value attained for these scenarios. 
Euclidean distance for scenario 1 is 0.63 because, although the prices of these two products are numerically different from each other, their normalized distances are relatively close (see Eqn.~\eqref{eq:customDistance}). 

\begin{table}[!ht]
    \centering
     \caption{Comparison of the distance metrics based on three different scenarios}
     \label{tab:scen_table}
     
   \begin{tabular}{lccc}
    \toprule
        Distance Metric & scenario 1 & scenario 2 & scenario 3\\
        \midrule
         MPBD & 0.00 & 0.10 & 0.60 \\
         Levenshtein & 1.00 & 1.00 & 0.30 \\
         Euclidean & 0.63 & 0.58 & 0.35\\
         \bottomrule
    \end{tabular}
    
\end{table}

Scenarios 1 and 2 are very similar, but in scenario 2, product0 price drops by one unit (i.e., string distance between B and A) on day 4, while product1 price drops by two units, and, on day 7, the product0 price increases by one unit and product1 price increases by two units. Although they move in exactly the same direction during the 10-day period, the magnitudes of the price movements are different from each other on the 4th and 7th days.
According to the MPBD, they are still very close to each other, however, since MPBD also considers the magnitudes of the movements, it results in a distance value of 0.1. 
When compared to scenario 1, since prices get closer numerically on days 4 and 7, the Euclidean distance becomes 0.58 for scenario 2. 

In scenario 3, the product prices move in opposite directions on the 3rd, 6th, and 8th days. Except for these three days, prices move steadily and show the same movement pattern. 
Besides, for a 5-day period, the prices continue at the same level. 
Accordingly, when measured with the Levenshtein and Euclidean distance measures, the price difference between these two products seems to be very small, while the MPBD takes the highest value among these three scenarios. 
This is because 3 out of 10 days, the prices of these two products move in opposite directions.

\subsection{Performance of the Clustering Algorithms}
\begin{figure*}[!ht]
    \centering
    \subfloat[Price dataset - CH index]{
    \includegraphics[width=0.45\textwidth]{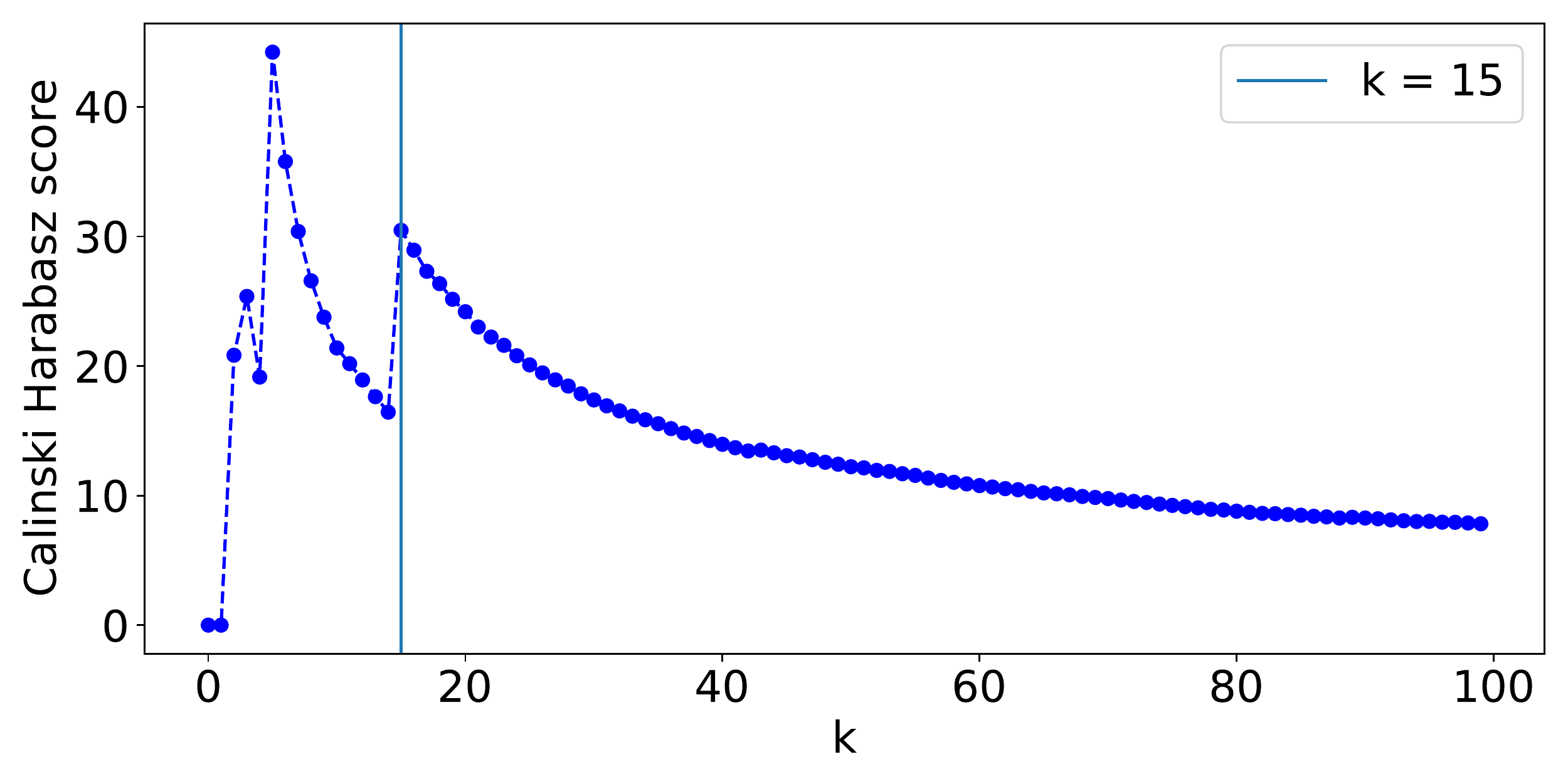}}
    \subfloat[Price dataset - DB index]{
    \includegraphics[width=0.45\textwidth]{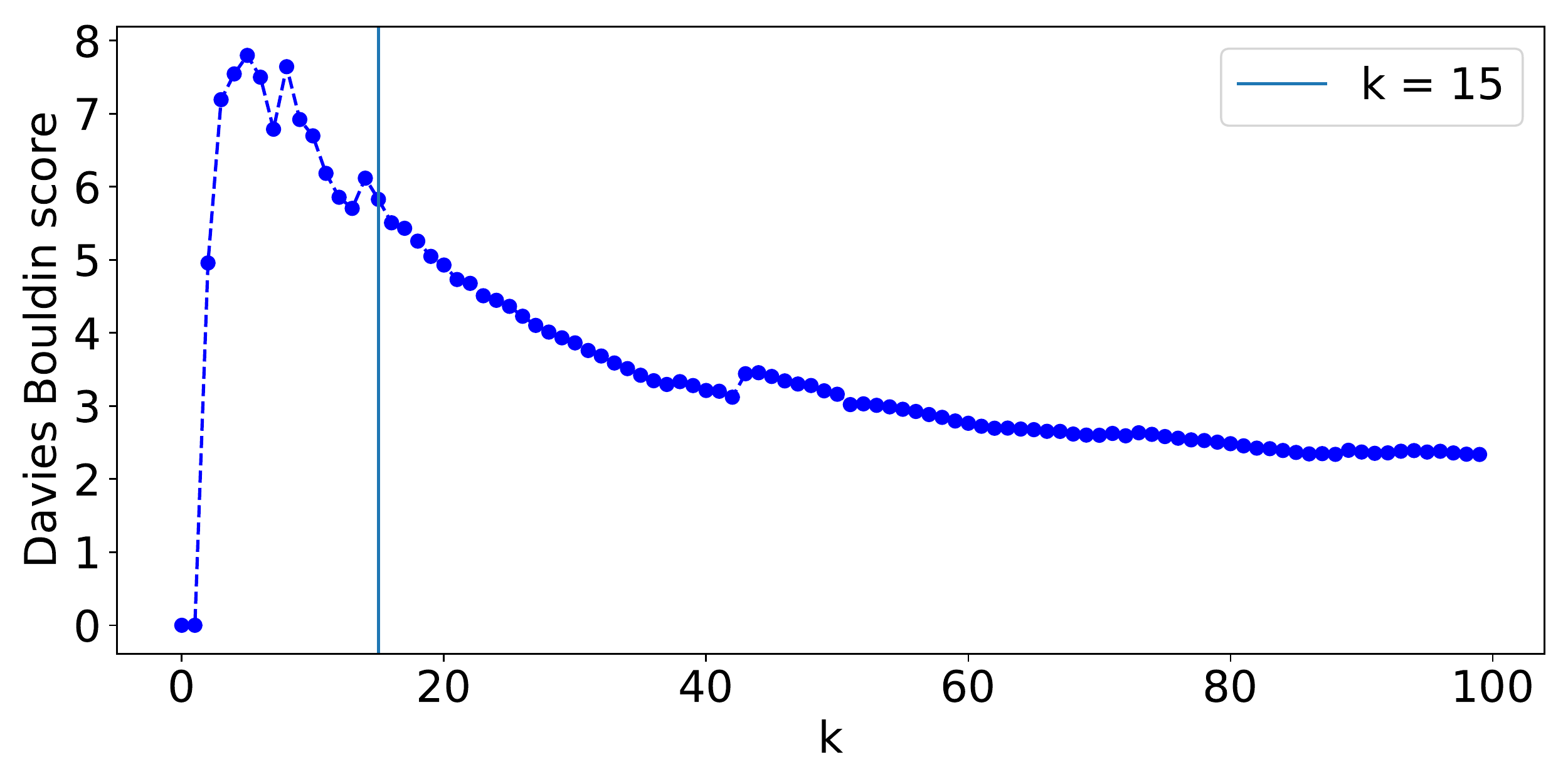}}\\[0.4em]
     \subfloat[Sales dataset - CH index]{
    \includegraphics[width=0.45\textwidth]{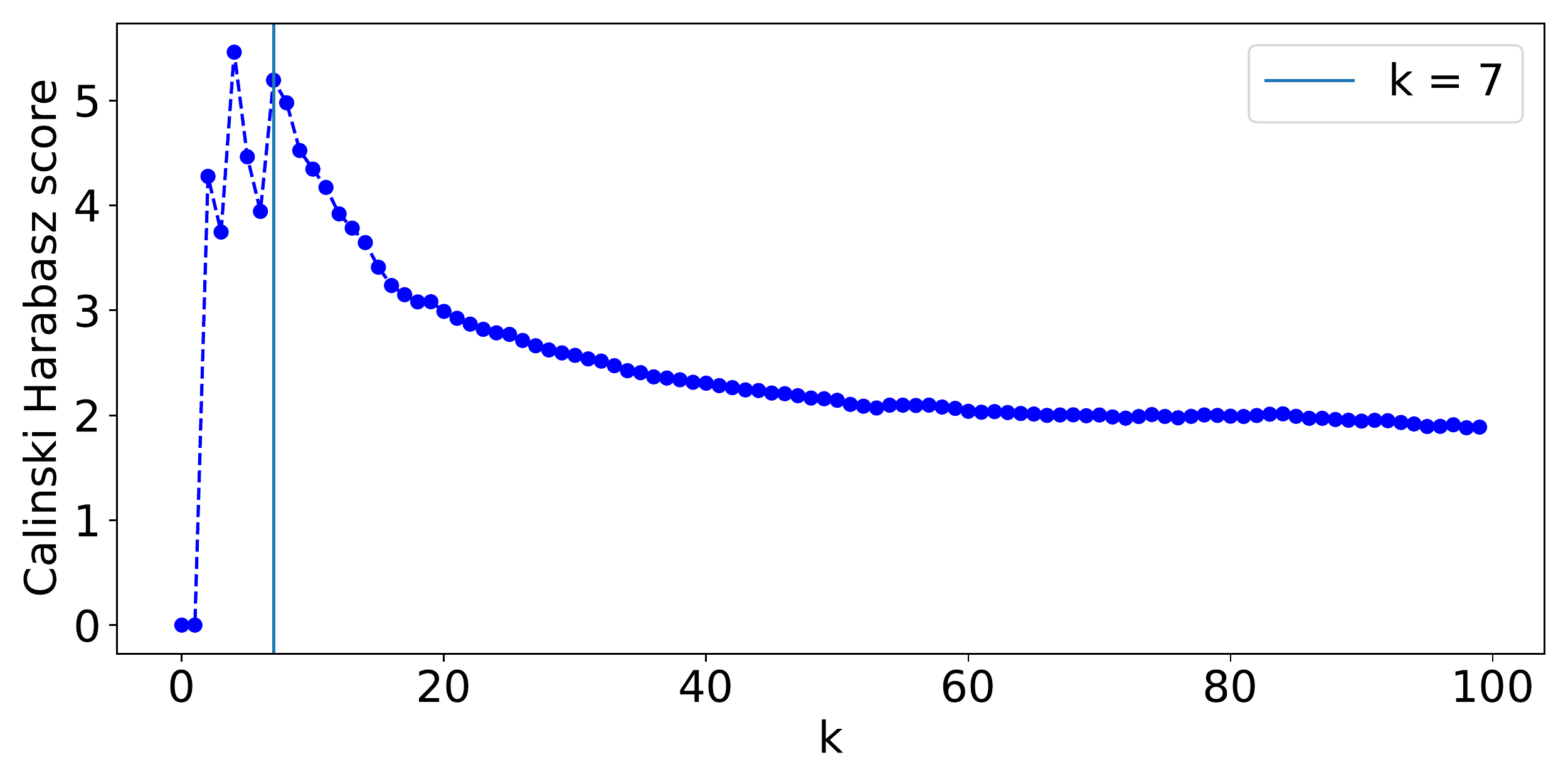}}
    \subfloat[Sales dataset - DB index]{
    \includegraphics[width=0.45\textwidth]{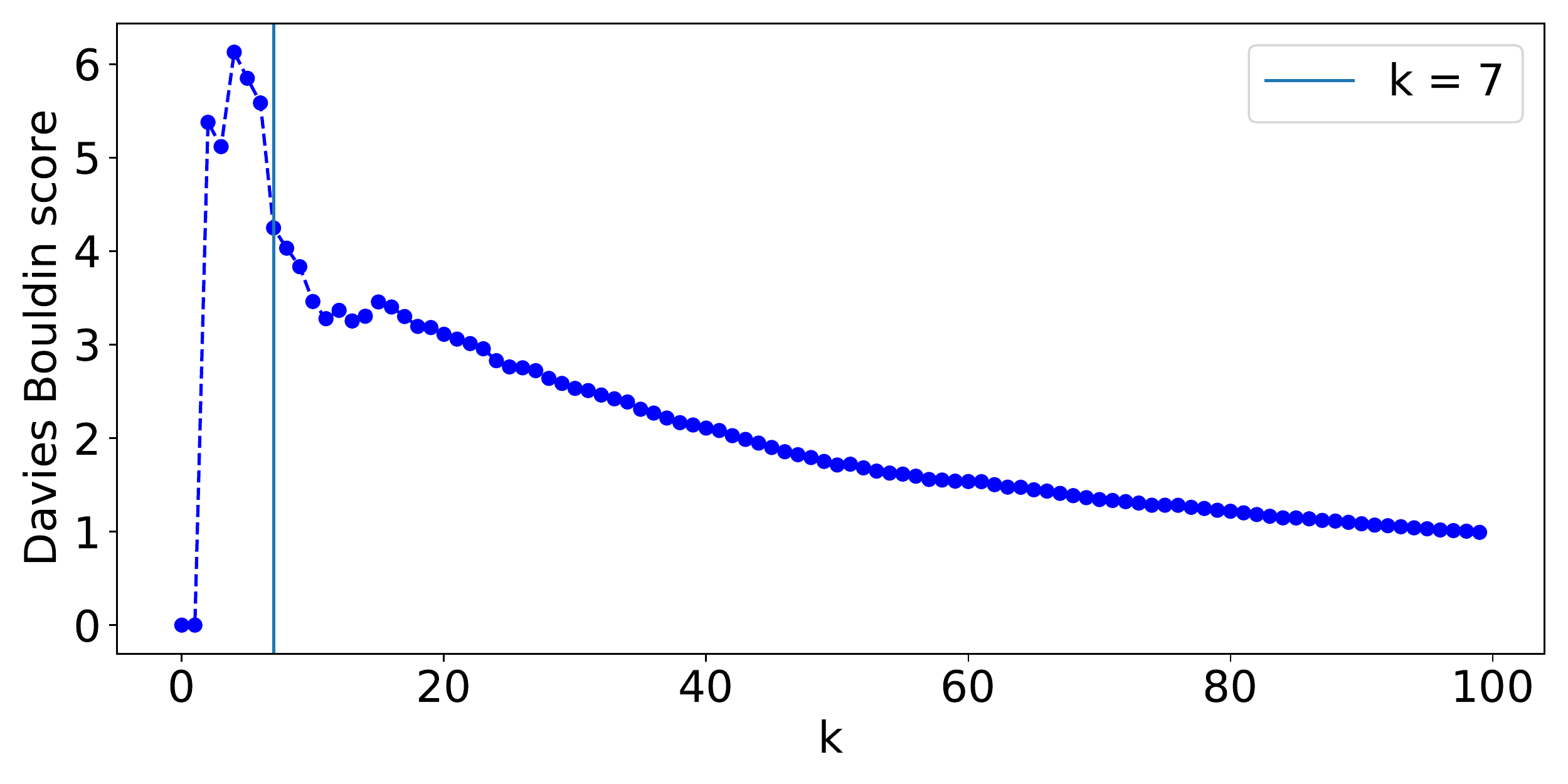}}
    \caption{Results of hierarchical clustering with MPBD}
    \label{fig:custom_results}
\end{figure*}

We evaluate the performances of hierarchical and $k$-means clustering algorithms with the combination of different distance metrics by using CH, DB, and the MBPI. 
For these experiments, we use normalized versions of the distance values generated by the different distance metrics to avoid any bias due to the scale of the metrics.
We first run the hierarchical clustering algorithm with the MPBD for a different number of clusters ranging from 0 to 100, and plot the graph of CH and DB scores (see Figure~\ref{fig:custom_results}). 
After 15 clusters, the CH score decreases significantly \tcolR{for the price dataset}. 
Although one of the lowest DB values seems to be obtained with 13 clusters, the most appropriate number of clusters is chosen as 15 \tcolR{for the price dataset}, since a very similar value was obtained with 15 clusters. 
\tcolR{For the sales dataset, after 7 clusters, the CH score decreases significantly. 
Although DB has lower values for a higher number of clusters, at 7 clusters, there is a sharp drop in its value.}
As a result, we consider $k=15$ clusters \tcolR{for price dataset and $k=7$ clusters for sales dataset} in the remaining experiments. 

Table~\ref{tab:results} presents the performances of $k$-means and hierarchical clustering algorithms with respect to DB, CH, and MBPI. 
\tcolR{For the price dataset,} we obtain the highest CH and the lowest DB score by using a $k$-means algorithm with the Euclidean distance. 
This is expected because the evaluation criterion for these two scores is the Euclidean distance, and they consider the numerical closeness of the product prices.
This algorithm is followed by a hierarchical clustering algorithm using Euclidean distance and Levenshtein distance. 
However, when we look at the MBPI, we see that the hierarchical clustering with the MPBD and image clustering ($k$-means with CNN-based feature learning) algorithms perform very similarly, and MBPI values are lower compared to other algorithms. 
That is, both hierarchical clustering with the MPBD and image clustering techniques provide better clusters than other algorithms based on the MBPI, as they both aim to find price change pattern similarities of products (i.e., time series), and generate the clusters accordingly.
\tcolR{For the sales dataset, the image clustering algorithm, which uses the CNN-based algorithm for feature extraction and $k$-means for clustering, obtains the best values for the DB and CH evaluation metrics, while $k$-means with Euclidean distance obtains the best value for the MBPI metric. 
Our clustering approach obtains comparable results with the image clustering approach in terms of the MBPI metric.}
\setlength{\tabcolsep}{6pt} 
\renewcommand{\arraystretch}{1.13} 
\begin{table*}[!ht]
    \centering
    \caption{Summary results for the clustering algorithms and distance metrics}
    \label{tab:results}
    \resizebox{0.8\textwidth}{!}{%
    \begin{tabular}{llrrrrrr}
    \toprule
     &  & \multicolumn{3}{c}{Price Dataset} & \multicolumn{3}{c}{Sales Dataset} \\
     \cmidrule(lr){3-5}\cmidrule(lr){6-8}
        Clustering Alg. & Distance Metric & DB & CH & MBPI & DB & CH & MBPI \\
        \midrule
        $k$-means & Euclidean &\textbf{1.7} & \textbf{198.4} & 1525.3&2.6 & 8.2 & \textbf{7046.5}\\
        $k$-means & CNN-based & 4.1 & 63.5 & 1494.7& \textbf{2.3} & \textbf{13.7} & 7394.8\\
        $k$-means & DTW &6.4 & 49.2 & 2259.0& 4.1 & 6.2 & 7481.6\\
        Hierarchical & Euclidean & 2.8 & 178.8 & 1551.5& 3.9 & 7.2 & 7175.3\\
        Hierarchical & Levenshtein & 3.8 & 149.2 & 1557.1& 4.1 & 6.7 & 7362.5\\
        Hierarchical & DTW &51.8  &4.7  &2269.0 & 6.1 &3.6  &7406.2 \\
        Hierarchical & MPBD & 5.8 & 30.4 & \textbf{1415.0} & 4.3 & 5.2 & 7125.0\\
        \bottomrule
    \end{tabular}}
\end{table*}

\tcolR{For the price dataset,} the number of time series in the clusters that are generated by both image clustering and the custom clustering approaches are not as homogeneous as the distribution of time series among the clusters generated by other algorithms. 
In this dataset, there are some products whose prices do not change or rarely change over the given time period. 
Because the price values are more stable compared to sales values, number of such products can be quite large and their prices typically differ from each other.
Although other traditional clustering techniques cannot capture such patterns, both image clustering and our custom clustering approach captures this pattern, and place these products in the same cluster. 
Therefore, in the clusters that are created by both image clustering and custom clustering, there are two clusters having more than 200 time series consisting of these products. 
\tcolR{Since the sales values fluctuate at a higher rate, it is more difficult to capture above mentioned pattern. 
However, we find that the image clustering approach generates very homogeneous clusters in terms of the number of time series for the sales dataset.
}

\begin{figure*}[!ht]
\centering
    \subfloat[Cluster 1 (9 time series)]{
    \includegraphics[width=0.45\textwidth]{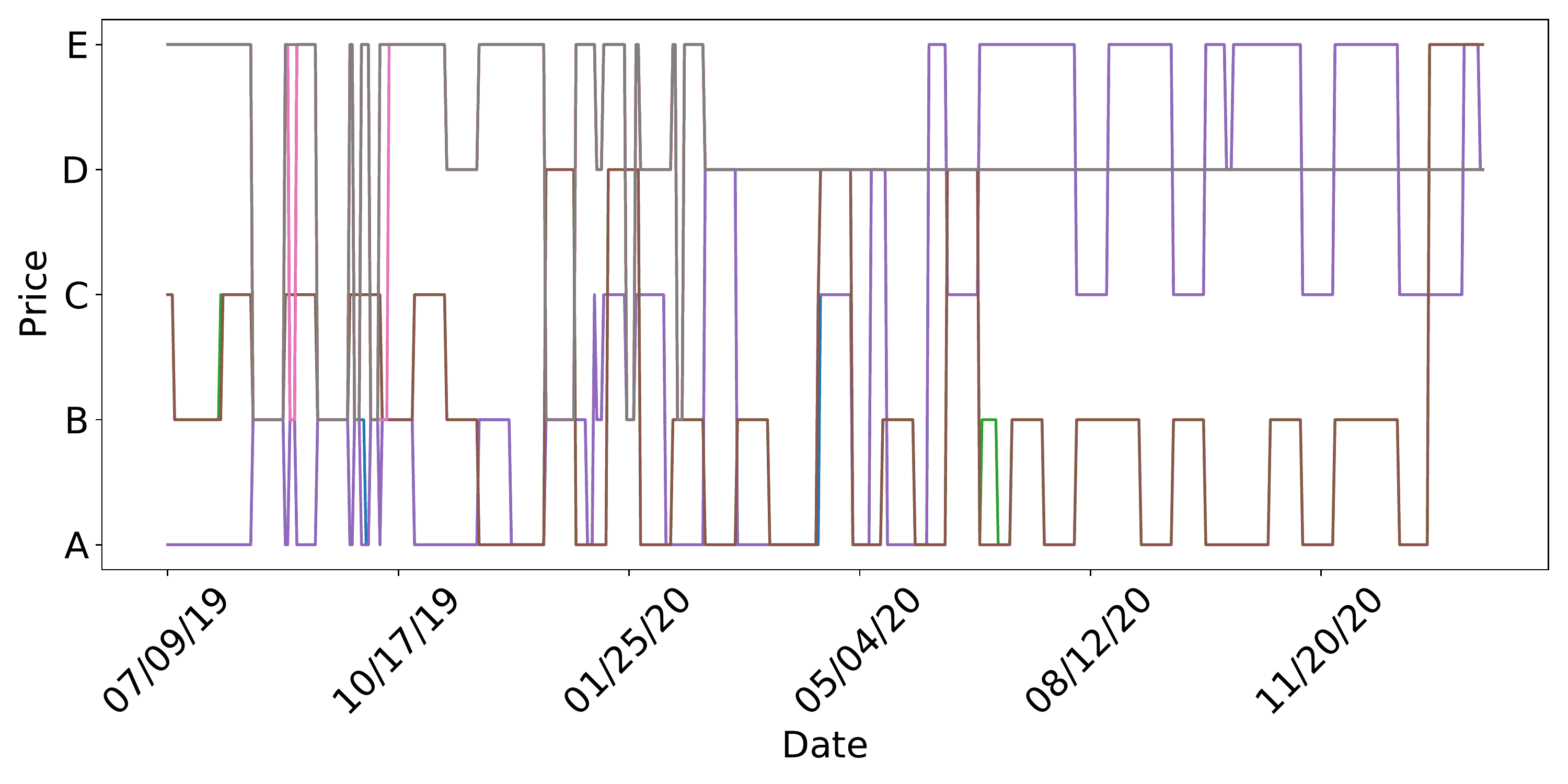}}
    \subfloat[Cluster 5 (68 time series)]{
    \includegraphics[width=0.45\textwidth]{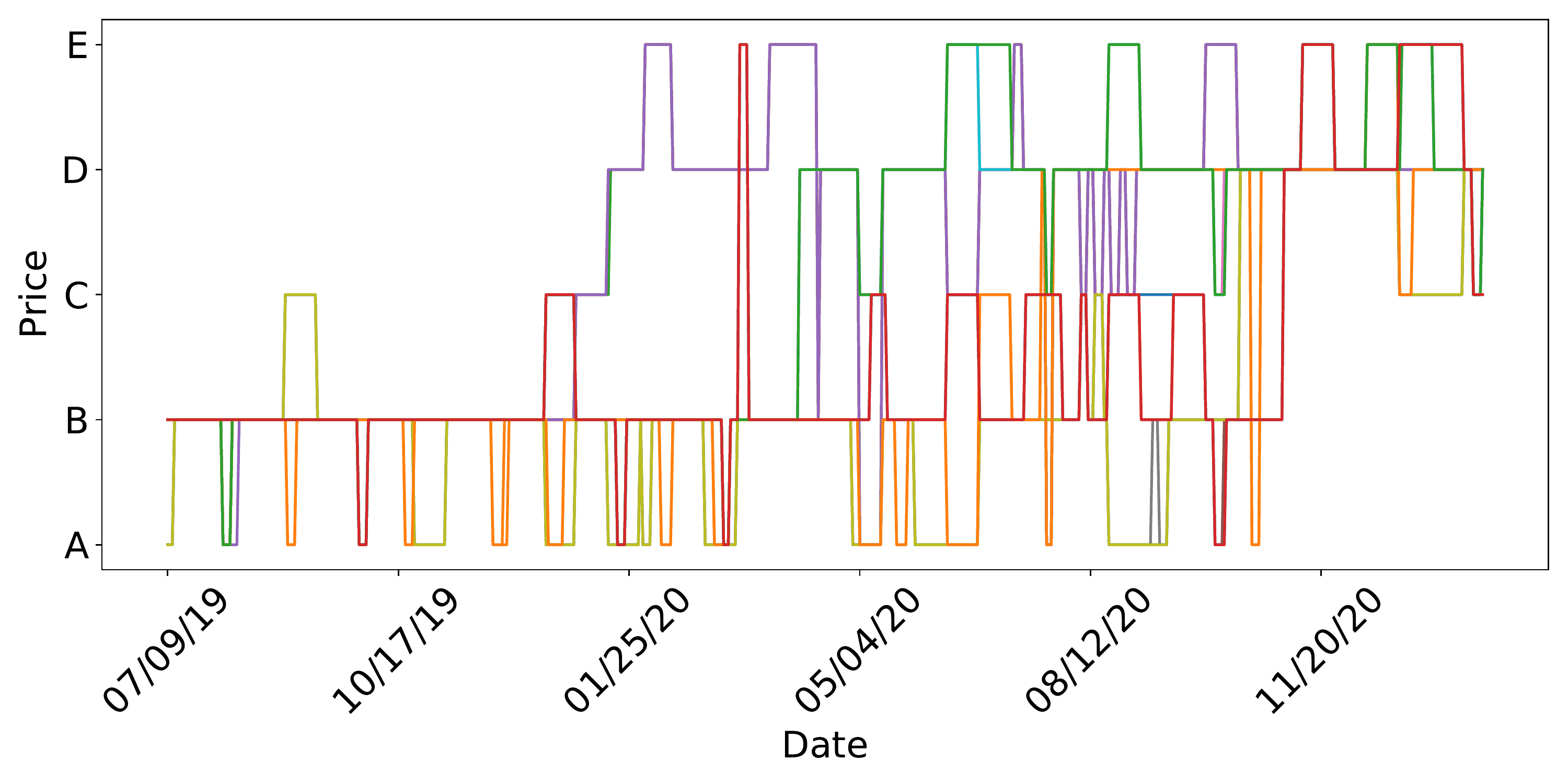}}
    \caption{ Two sample clusters generated by the hierarchical clustering with MPBD for the price dataset.}
    \label{fig:custom_cluster_samples}
\end{figure*}

Figure~\ref{fig:custom_cluster_samples} presents time series in two sample clusters generated by the custom clustering approach for the price dataset. 
As well as the custom clustering approach captures the time series showing almost the same pattern in the same price level, it also captures the similar patterns that occur in different price levels. 
Both in clusters 1 and 5, although there are some overlapping patterns in the same level (especially in cluster 5), there are similar patterns in different price levels.

\begin{figure*}[!ht]
\centering
    \subfloat[Cluster 3 (22 time series)]{
    \includegraphics[width=0.90\textwidth]{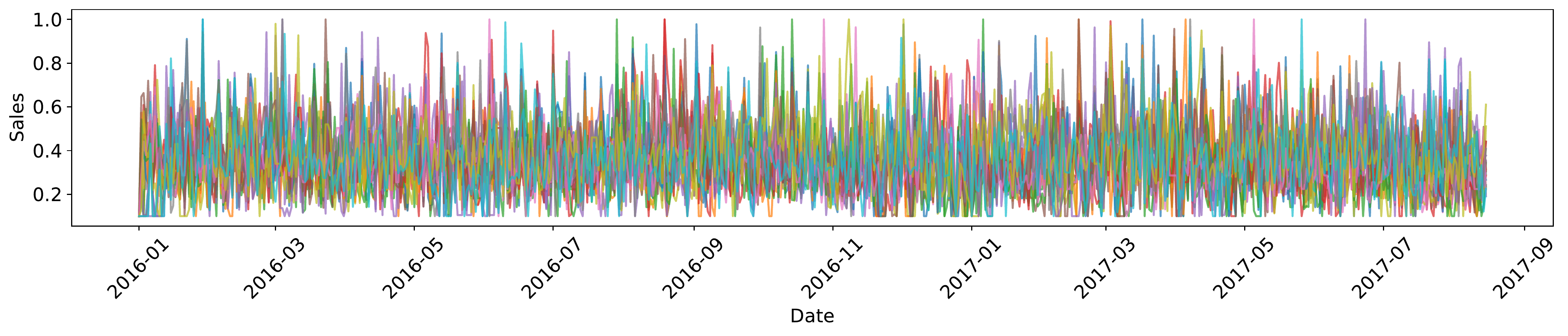}}\\
    \subfloat[Cluster 5 (16 time series)]{
    \includegraphics[width=0.90\textwidth]{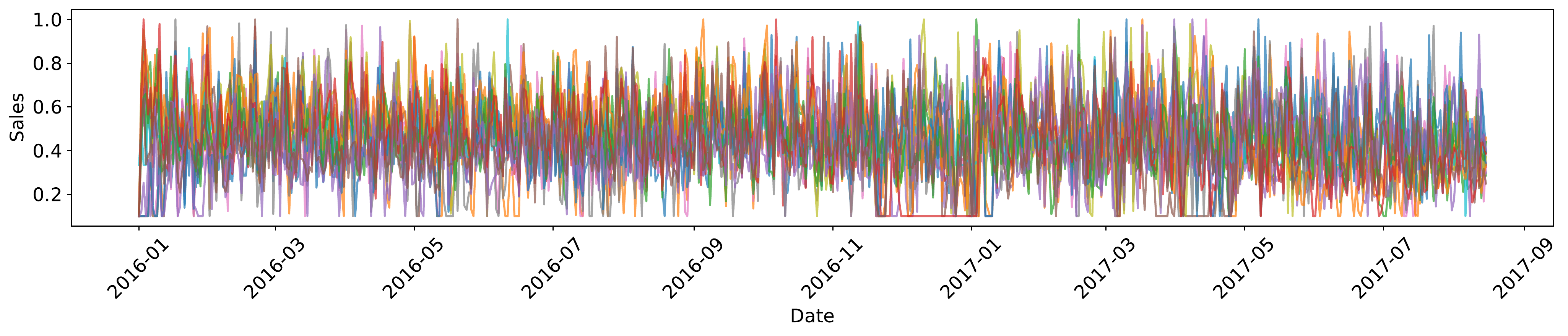}}
    \caption{ Two sample clusters generated by $k$-means clustering with Euclidean distance for the sales dataset.}
    \label{fig:kmeans_euclid_cluster_samples}
\end{figure*}

Figure~\ref{fig:kmeans_euclid_cluster_samples} presents two sample clusters generated using the $k$-means algorithm with Euclidean distance for the sales dataset. 
We note that this approach results in time series clusters that are close together spatially, but may not capture the similarity of the underlying movement patterns.

\subsection{Cluster Profiling}
\tcolR{We create the cluster profiles based on the available features of both datasets to better understand the cluster contents. Table~\ref{tab:profiling_price} presents the cluster-ID, the number of time series, the number of product categories, top two categories having the highest number of products, average, min, and max prices of each cluster that are generated by hierarchical clustering with MPBD metric using the price dataset. The minimum number of categories in clusters is one and the maximum number of categories is 15. However, the top two categories cover almost 50\% of all products in each cluster. The average product prices in the clusters vary between 9 and 34, while the minimum prices of the products are between 1 and 10, and finally, the maximum price values change between 6 and 120. Some products in the Personal Care category and the entire Pet category are separated from other products, and form homogeneous clusters. 
Other categories that are often in the same cluster are (Snacks, Water \& Beverage), (Snacks, Essential Food) and (Essential Food, Water \& Beverage). }
\setlength{\tabcolsep}{3pt} 
\renewcommand{\arraystretch}{0.93} 
\begin{table*}[!ht]
 \centering
    \caption{Cluster profiling for the price dataset}
    \label{tab:profiling_price}
    \resizebox{\textwidth}{!}{%
\begin{tabular}{lrclrrr}
\toprule
cluster & size & \# of categories & top 2 categories (\# of products) & avg. price & min price & max price \\
\midrule
1 & 9 & 1 & Personal Care: 9 & 34 & 10  & 80  \\
\midrule
2 & 8 & 1 & Personal Care: 8 & 18 & 8 & 26  \\
\midrule
3 & 9 & 1 & Pet: 9 & 4  & 3   & 6   \\
\midrule
4 & 20 & 4 & Home \& Life: 7 & 14 & 1 & 30  \\
&   &   & Home Maintenance: 7 & &   &\\
\midrule
5   & 68   & 8 & Personal Care: 18 & 19 & 2 & 63  \\
&   &   & Water \& Beverages: 15 &  &   &\\
\midrule
6   & 23    & 2 & Home Maintenance: 5 & 19 & 2   & 42\\
&   &   & Personal Care: 18 &    &     &\\
\midrule
7   & 56    & 10 & Personal Care: 12    & 18 & 3   & 120 \\
&   &    & Home \& Life: 10 &   &   &\\
\midrule
8   & 82   & 8  & Essential Food: 21    & 14 & 1   & 78\\
&   &    & Personal Care: 20    &   &   &\\
\midrule
9   & 22    & 6 & Baby: 6    & 15 & 4   & 40\\
&   &   & Water \& Beverages: 5 &   &   &\\
\midrule
10  & 69    & 8 & Essential Food: 27    & 9  & 2   & 37\\
&   &    & Water \& Beverages: 12, Baby: 12 &    &     &\\
\midrule
11  & 71   & 11 & Snacks: 20    & 11 & 1    & 40\\
&   &   & Essential Food: 15    &   &   &\\
\midrule
12  & 261  & 12 & Water \& Beverages: 57 & 11 & 1   & 110\\
&   &   & Snacks: 55    &    &     &\\
\midrule
13  & 234  & 15 & Essential Food: 46, Milk and Breakfast: 40 & 12 & 1 & 90  \\
&   &    & Snacks: 47   &    &     &\\
\midrule
14  & 114  & 13 & Essential Food: 30 & 12 & 1   & 87\\
&   &   & Water \& Beverages: 23    &    &     &\\
\midrule
15  & 80    & 10    & Essential Food: 22    & 13 & 1   & 55\\
\bottomrule
\end{tabular}}
\end{table*}

When the clusters in Table~\ref{tab:profiling_price} are examined closely, we find that the products whose prices do not change or rarely change are ice creams, books and magazines, frozen and canned goods, nuts, sauces and condiments, and jams. 
Another interesting finding is that clusters generated by standard clustering algorithms and distance measures do not reveal the relationships between product categories and the price change patterns. 
However, image clustering and hierarchical clustering with MPBD tend to generate clusters that are consisting of a single product category, such as personal care and pet products. 
In addition, we observe that the products that are likely to be sold together (e.g., cereal, avocado, bowl) show similar price change patterns, and are grouped under the same cluster.

\tcolR{Table~\ref{tab:profiling_sales} presents the similar statistics to those of Table~\ref{tab:profiling_price} for the sales dataset, which are obtained by using $k$-means clustering approach (i.e., the best performing clustering method for this dataset).
Note that this dataset does not include detailed information (e.g., product category names), thereby limiting this analysis. 
The number of products in each cluster is fewer than the number of product-store sales time series. 
This means that the same product even if it is sold in different stores shows a similar sales pattern. 
Since the number of categories in the data set after preprocessing step is very small, we cannot make a detailed inference as we did for the price data set. 
We observe that cluster four is the cluster with high sales, while clusters one, three, and seven are low-sales clusters, and clusters two, five and six are moderate-level sales clusters.}



\setlength{\tabcolsep}{3pt} 
\renewcommand{\arraystretch}{0.93} 
\begin{table*}[!ht]
\centering
\caption{Cluster profiling for the sales dataset}
\label{tab:profiling_sales}
\resizebox{\textwidth}{!}{%
\begin{tabular}{lrrrrrrr}
\toprule
cluster & size & \# of products & \# of stores & \# of categories & avg. sales & min sales & max sales \\
\midrule
1       & 49   & 24              & 9           & 1                & 5         & 0         & 53        \\
2       & 22   & 13             & 5          & 1  & 9         & 0         & 80        \\
3       &22   & 15              & 7           & 2                & 5        & 0         & 58       \\
4       & 19   & 14  & 3          & 1                & 19         & 0         & 109        \\
5       & 16   & 15             & 5           & 2                & 9         & 0         & 67        \\
6       & 5   & 8              & 1           & 1                & 11        & 0         & 39        \\
7       & 1   & 1              & 1           & 1                & 4        & 0         & 21      \\
\bottomrule
\end{tabular}}
\end{table*}

\section{Conclusion}\label{sec:conc}
In this study, we focus on clustering grocery products based on their price and sales volume histories, which can be represented as time series.
For the price dataset, this analysis can be used to gather insights about the products having price hikes and discounts in the same periods. 
On the other hand, for the sales dataset, finding the products with similar sales patterns can be used for various marketing operations and analysis, including which products to use in promotions and market basket analysis.
We develop a custom distance (MPBD) and also a custom evaluation metric (MBPI) for this purpose. 
We also employ the image clustering technique to group the products based on their price change patterns. 
We compare these approaches with the traditional clustering methods and evaluate their performances by using CH, DB, and the MBPI metrics.
The results show that the hierarchical clustering with the MPBD and the image clustering algorithm achieve comparable results on the time series product price dataset, and outperform traditional techniques, whereas high level of fluctuations and low number of time series in the sales dataset largely prevent having a dominant clustering technique. 
Overall, we find both our proposed method and the image clustering as suitable approaches for clustering over grocery product datasets to identify similar time series based on their movement patterns. 
The work presented in this paper can be further extended by applying both custom and image clustering approaches on different time series datasets.
These clustering approaches could also benefit from application to similar datasets and tasks to strengthen our findings. 
Furthermore, we note that many global forecasting models may be able to successfully leverage the clustering data generated by the proposed methods in this study, e.g., for the purposes of increasing predictive power for forecasting.

\section*{Acknowledgements}
The authors would like to thank the 
XYZ 
company for supporting this study and providing data and feedback throughout.

\section*{Data availability statement}
The Favorita dataset that supports the findings of this study is available at the following link: \\ 
\url{https://www.kaggle.com/c/favorita-grocery-sales-forecasting/data}\\
Price dataset is propriety and obtained from 
XYZ company.

\section*{Disclosure statement}
No potential conflict of interest was reported by the authors.

\bibliographystyle{plainnat} 
\bibliography{bib}
\end{document}